%% file: HumynexSurg1.tex
\documentclass[11pt]{article}
\usepackage[margin=1in]{geometry}
\usepackage{graphicx}
\usepackage{booktabs}
\usepackage{amsmath}
\usepackage{amssymb}
\usepackage{xcolor}
\usepackage{siunitx}
\usepackage[colorlinks=true,linkcolor=blue!50!black,citecolor=blue!50!black,urlcolor=blue!50!black]{hyperref}
\usepackage{caption}
\usepackage{titlesec}
\usepackage{tikz}
\usepackage{float}
\usepackage{array}
\newcolumntype{R}[1]{>{\raggedright\arraybackslash}p{#1}}
\usetikzlibrary{arrows.meta,positioning,fit,calc}
\titlespacing*{\section}{0pt}{1.1em}{0.5em}
\titlespacing*{\subsection}{0pt}{0.9em}{0.4em}
\newcommand{\dsname}{HumynexSurg-1}
\newcommand{\good}{\textcolor{green!45!black}{\checkmark}}
\newcommand{\bad}{\textcolor{red!70!black}{$\times$}}
\input{v3_numbers.tex}

\input{v3_scaling_numbers.tex}

\title{{\Large\bfseries \dsname{}: A Curated Expert Liposuction Dataset}\\[2pt]
\large Synchronized telemetry, video and think-aloud narration from a master surgeon,
packaged for robot foundation models}
\author{Rhea Huang$^{1}$ \qquad David L.\ Matlock, MD, MBA$^{1}$ \qquad Dr.\ Laurence Reich$^{1}$\\[2pt]
\normalsize $^{1}$Humynex Robotics, Inc., Beverly Hills, CA\\
\normalsize Correspondence: \texttt{rhuang@humynex.ai} \,/\, \texttt{drmatlock@humynex.ai} \,/\, \texttt{larmd@humynex.ai}}
\date{September 2026 --- version 0}

\begin{document}
\maketitle
\vspace{-1.2em}

\begin{abstract}
\noindent
Robot foundation models learn manipulation from large demonstration corpora, but surgery
is missing from those corpora: across the 780-hour Open-H surgical collection, one dataset
carries synchronized force and none covers an aesthetic procedure. Liposuction is the hard
case, because the instrument works under the skin and the surgeon operates by feel and by
judgment. Humynex Robotics builds curated expert datasets for
this kind of procedure. \dsname{} is the first release: a master liposuction surgeon
performing on porcine abdominal tissue while narrating every decision, recorded with
synchronized suction pressure, six-axis hand force/torque, top-down RGB-D video, side video
and a lavalier microphone --- 14 episodes, 42{,}738 frames, 35.6 minutes, 356 utterances of
which 95\% compile into a liposuction-specific label schema. The capture follows a patent-pending sensing plan
organized around the quantities a policy needs, so a
channel captured today by a model can be upgraded to a sensor tomorrow without changing the
data format. This release captures the instrument motion as a tool-hand track in the side video and
provides the force channel as state; the funded capture adds a measured 6-DoF handle pose,
a validated force channel, ultrasound imaging of the fat layer, and palpation sensing. As a
proof of concept, NVIDIA Isaac GR00T N1.7 fine-tunes on the dataset with no custom code in
under an hour per run and learns the recorded sessions; scaling probes on the same episodes
show where further gains come from: every new session lowers the error on an unseen
session. The dataset, its label schema, its quality-assurance reports and its evaluation
protocol are the product; the next capture, many short
sessions across fat regions with the sensors named here, is what the probes point to.
\end{abstract}

\section{Why a curated expert dataset}
\label{sec:why}

\textbf{The gap.} Foundation policies such as GR00T~\cite{gr00t17}, ACT~\cite{zhao2023act},
Diffusion Policy~\cite{chi2023diffusion} and $\pi_0$~\cite{black2024pi0} learn from
demonstrations. Surgical variants exist: GR00T-H~\cite{gr00th} post-trains on 601 hours of
surgical robot data and SRT-H~\cite{kim2025srth} reached phase-level autonomy from narrated
demonstrations. The large corpora are video plus kinematics. They carry little force, no
suction, and no record of what the surgeon was thinking. GR00T-H's own authors report that
the model is weakest on contact-rich steps and ask for tissue-contact labels~\cite{openh}.

\textbf{Why liposuction.} The cannula tip works out of sight under the skin. The surgeon
steers by resistance, by the sound of the suction line, by palpating with the other hand,
and by a decision vocabulary (plane, tract, pinch test) that is spoken in the operating room
and recorded nowhere. A camera alone cannot capture this skill; a dataset that captures it
must be designed around the quantities the skill depends on.

\textbf{Why now.} Open-H Embodiment~\cite{openh} standardized how surgical demonstration data
is packaged and showed who the buyers are: model builders with robots and compute but no
procedure data. Exactly one of its 119 datasets has synchronized force and none touches
aesthetic surgery. A force-and-narration liposuction corpus, in the ecosystem's own format,
fills a verifiable hole.

\textbf{What this document is.} \dsname{} version~0 is the first release of a curated
expert corpus. Sections~\ref{sec:expert}--\ref{sec:dataset} describe the expert, the sensing
plan and the data. Section~\ref{sec:quality} describes the quality assurance that ships with
the data. Section~\ref{sec:poc} is a proof of concept of what a buyer
can do with the data. Section~\ref{sec:next} states what the next capture adds and what it
costs.

\section{The expert and the tissue}
\label{sec:expert}

\subsection{Expert demonstrator}
\begin{figure}[t]
\centering
\includegraphics[width=\linewidth]{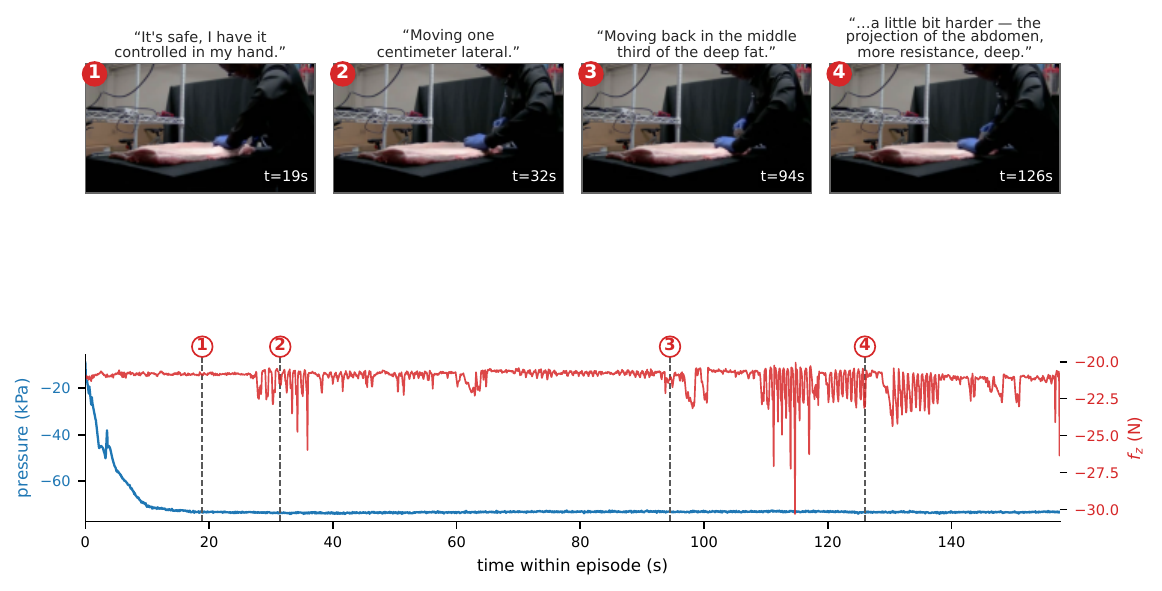}
\caption{One demonstration at a glance (episode~0): side-camera frames with the surgeon's
simultaneous narration (numbered markers) above the synchronized suction-pressure (blue) and
force (red) telemetry.}
\label{fig:storyboard}
\end{figure}

\textbf{David L.\ Matlock, MD, MBA, FACOG}, CEO and co-founder of Humynex Robotics.
\begin{itemize}\setlength\itemsep{0.05em}
  \item About four decades of cosmetic-surgery practice; an operator-reported career volume
  above 10{,}000 liposuction procedures (1987--2026);
  \item Published on ultrasound-guided fat grafting, including a sonographic study of
  autologous fat transfer~\cite{matlock2014};
  \item Narrates in a stable operative vocabulary (planes, tracts, resistance, pinch test),
  which is what lets the narration compile into labels deterministically (Sec.~\ref{sec:labels}).
\end{itemize}
\textbf{Why one expert.} Liposuction skill is dominated by tactile judgment that leaves no
trace in video. A single-operator corpus trades population diversity for consistency: every
episode reflects one coherent decision policy, which is the right regime for asking whether
expert demonstrations carry learnable structure at all. Inter-surgeon variation is a later
release. Figure~\ref{fig:storyboard} shows the raw material.

\subsection{Tissue and sessions}
Version~0 uses porcine abdominal wall with skin, fat and muscle intact. It is commercially
available, ethically simple and mechanically a mammalian subcutaneous fat system, and it can
be worked at full force. A cadaveric-tissue phase is planned to ground anatomical fidelity, with the force and pressure profiles captured here as the
transfer baseline.

The three sessions worked different regions of the specimen, and the narration shows that those regions were different material
(Table~\ref{tab:tissue}): session~1 reads as fibrous, deep and hard; session~2 as firm,
superficial and high-resistance; session~3 as yielding, low-resistance and easy. This is the
variation a policy must learn across, and the scaling probes of Section~\ref{sec:poc}
single it out as the lever for further gains. It also sets the design of the next capture:
one session per fat region and composition (Sec.~\ref{sec:next}).

\input{tables_v4_tissue.tex}

\section{What we capture: the sensing plan}
\label{sec:sensing}

\begin{figure}[t]
\centering
\resizebox{\linewidth}{!}{%
\begin{tikzpicture}[
  font=\scriptsize, align=center, node distance=3mm,
  ch/.style={draw, rounded corners=1.5pt, inner sep=3pt, fill=blue!5, minimum height=8mm, text width=36mm},
  q/.style={draw, rounded corners=1.5pt, inner sep=3pt, fill=green!8, minimum height=8mm, text width=27mm},
  st/.style={draw, rounded corners=1.5pt, inner sep=4pt, fill=orange!12, minimum height=10mm, text width=34mm},
  arr/.style={-{Stealth[length=1.6mm]}, semithick}]
\node[ch] (c1) {\textbf{instrument motion}\\v0: side video $\rightarrow$ \mbox{2-D} tracker (model)\\next: 6-DoF pose sensor on the handle};
\node[ch, below=of c1] (c2) {\textbf{tissue state}\\v0: not captured\\next: ultrasound imaging of the fat layer, per region};
\node[ch, below=of c2] (c3) {\textbf{contact}\\v0: hand force/torque, suction pressure\\next: validated force; contact and line sensing};
\node[ch, below=of c3] (c4) {\textbf{palpation}\\v0: video track of the other hand (model)\\next: tactile sensing on the other hand};
\node[ch, below=of c4] (c5) {\textbf{context}\\v0: top \mbox{RGB-D}, side camera\\next: triggered cameras + egocentric view};
\node[ch, below=of c5] (c6) {\textbf{cognition (narration)}\\v0: lavalier mic $\rightarrow$ ASR $\rightarrow$ tags\\next: timecoded audio};
\node[q] (q1) at (5.0,0 |- c1) {where the instrument is\\and how it moves};
\node[q] (q2) at (5.0,0 |- c2) {what the tissue under\\the skin is like};
\node[q] (q3) at (5.0,0 |- c3) {what the instrument touches\\and how hard};
\node[q] (q4) at (5.0,0 |- c4) {what the other hand feels};
\node[q] (q5) at (5.0,0 |- c5) {scene context, registration, QA};
\node[q] (q6) at (5.0,0 |- c6) {intent, sensation, assessment\\(\texttt{lipo\_v0} labels)};
\node[st] (s) at ($(10.0,0 |- c1)!0.5!(10.0,0 |- c6)$) {\textbf{machine-learning record}\\state $\cdot$ images $\cdot$ language\\$\rightarrow$ action (next-step motion)\\LeRobot / GR00T format};
\foreach \i in {1,...,6} {\draw[arr] (c\i) -- (q\i);}
\foreach \i in {1,...,6} {\draw[arr] (q\i.east) -- (s.west);}
\end{tikzpicture}}
\caption{The sensing plan (patent pending). Left: the sensing channels, with what
version~0 used and what the upgrade is. Middle: the quantity each channel supplies. Right:
the record a model consumes. A channel realized by a model today (the tracker) is upgraded
to a sensor tomorrow without changing the record.}
\label{fig:sensing}
\end{figure}
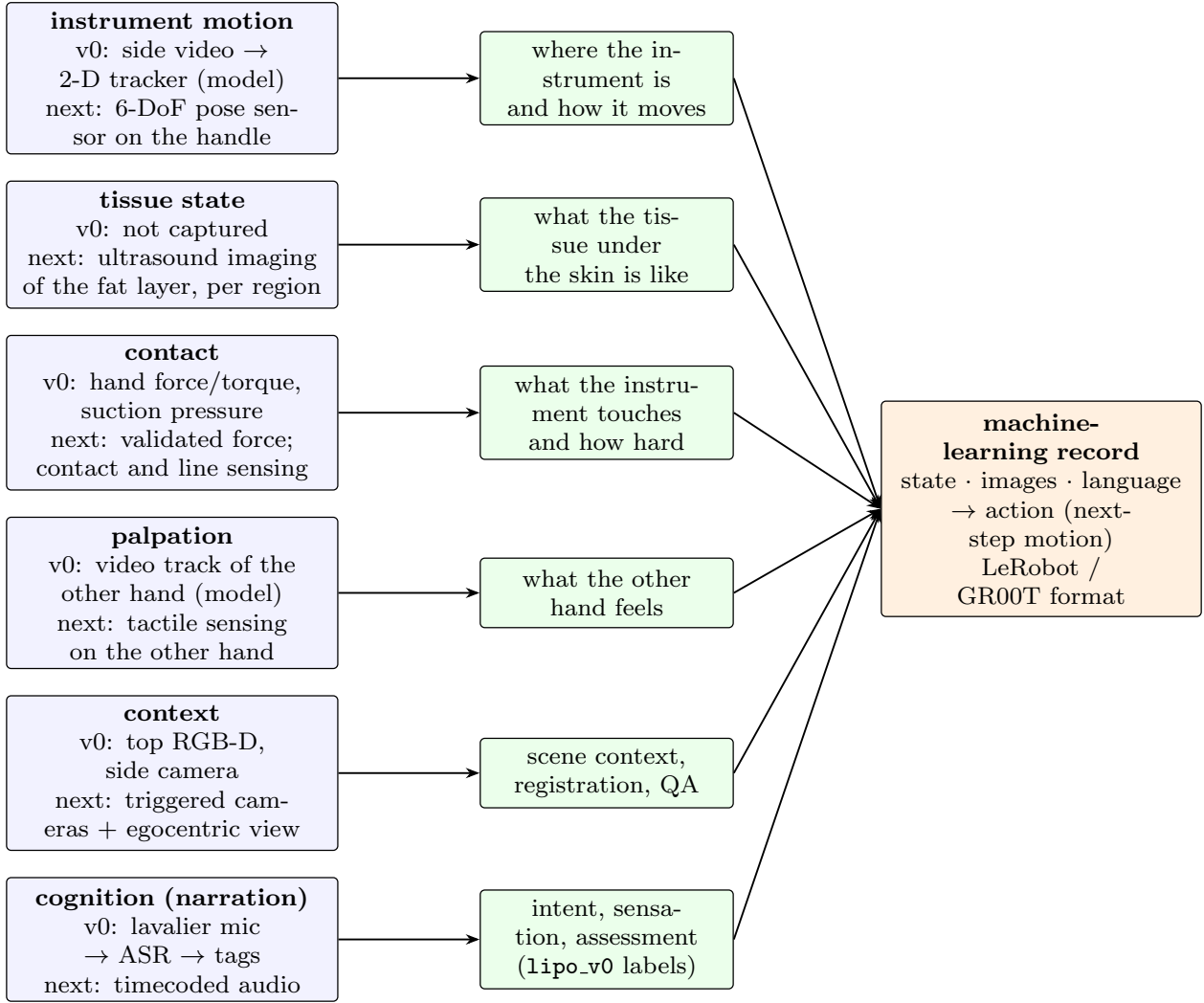

The plan is organized around the \emph{quantities} a policy needs (Fig.~\ref{fig:sensing})
and is the subject of two pending patent applications. A policy that drives a cannula needs to know how the instrument moves, what
the tissue under the skin is like, what the instrument is touching and how hard, what the
surgeon's other hand feels, what the scene looks like, and what the surgeon intends and
observes. Each quantity is filled by the best sensor available at the time of capture. A
quantity that is not yet sensed is estimated by a model from the channels that are, and is
flagged as such; when a sensor arrives, it upgrades the channel without changing the data
format, the state, or the action. Table~\ref{tab:sensing} lists, for every quantity, what
version~0 captured and how, and what the upgrade is.

\begin{table}[!t]
\centering\footnotesize
\caption{Quantity by quantity: why it matters for liposuction, what version~0 captured and
how, and what the funded sensor changes in the data. ``Model'' means the quantity was
estimated from other channels; ``sensor'' means it was measured. Costs are the
capture-system plan's tier totals: tier~A about \$2--4k on the rig we own, tier~B about
\$8--15k, tier~C about \$15--40k.}
\label{tab:sensing}
\begin{tabular}{R{2.0cm}R{3.3cm}R{3.8cm}R{5.5cm}}
\toprule
Quantity & Why it matters & Version 0 & Funded upgrade and what it changes in the data \\
\midrule
Instrument motion & the surgeon's motor output; the action a robot must reproduce &
\textbf{model}: tool-hand and support-hand positions tracked in the side video
(Sec.~\ref{sec:action}); 2-D, camera-relative; tool hand directly observed in
\vthreeTrackValidSOne{}--\vthreeTrackValidSThree{} of frames &
\textbf{sensor}: a 6-DoF pose sensor on the handle (tier A) gives position and orientation
including roll, in metric units, for every frame; ground-truth tracking for validation (tier C) \\
\addlinespace[5pt]
Tissue state & where the fat is and how thick; which layer the tip is working in &
not captured &
\textbf{sensor}: hand-held ultrasound imaging of the fat layer per region before, during and
after the procedure (tier B) adds the state of the tissue the tip is working in, which the
narration shows varies by region (Table~\ref{tab:tissue}) \\
\addlinespace[5pt]
Contact & resistance, layer transitions, line occlusion: what the surgeon feels &
\textbf{sensor}: 6-axis hand force/torque at 467\,Hz, provided as state
(Sec.~\ref{sec:quality}); suction pressure at 100\,Hz &
force/torque re-mounted and validated before every session (tier A) makes the force channel
a second action target; contact sensing at the instrument and flow sensing in the suction line
(tier B) add tissue-contact signals the surgeon hears and feels today \\
\addlinespace[5pt]
Palpation & the other hand's assessment of the tissue &
\textbf{model}: support-hand track from the side video &
\textbf{sensor}: tactile sensing on the other hand (tier B) records what the surgeon feels
when assessing the tissue, instead of only where that hand is \\
\addlinespace[5pt]
Context & scene, region registration, quality assurance &
\textbf{sensor}: top-down colour + metric depth; side camera; software timing &
hardware-triggered cameras on a shared timebase and an egocentric view (tier A) replace
software timing with per-frame synchronization \\
\addlinespace[5pt]
Cognition & intent, sensation, assessment: the decision layer &
\textbf{sensor}: lavalier narration $\rightarrow$ word-aligned transcript $\rightarrow$
\texttt{lipo\_v0} tags (95\% of utterances tagged) &
timecoded audio (tier A); reviewed label schema v1 \\
\bottomrule
\end{tabular}
\end{table}

\subsection{Hardware as fielded}
\begin{table}[h]
\centering\small
\caption{Capture hardware in version~0 (delivered rates).}
\label{tab:hw}
\begin{tabular}{lR{3.2cm}R{5.4cm}R{3.0cm}}
\toprule
Modality & Device & Delivered specification & Timestamping\\
\midrule
Top-down RGB-D & Intel RealSense D455 & colour $1280\times720$; depth $848\times480$, metric & device + host stamps\\
Side video & ELP USB camera & $480\times270$ at 20\,Hz in the dataset & host stamps\\
Hand force/torque & 6-axis force/torque sensor & 6-axis wrench, $\sim$467\,Hz delivered & host arrival stamps\\
Suction pressure & gauge + LabJack U3-HV & 100\,Hz analog & software-timed\\
Narration & wireless lavalier mic & 48\,kHz; WhisperX~\cite{bain2023whisperx} word times & aligned to host clock\\
\bottomrule
\end{tabular}
\end{table}

All streams share the host clock and are resampled to a 20\,Hz grid; raw cadences are
preserved and each stream ships with a timing report. The next capture adds hardware
triggering (Sec.~\ref{sec:next}).

\subsection{The action channel: tool-hand kinematics}
\label{sec:action}
\begin{figure}[t]
\centering
\includegraphics[width=\linewidth]{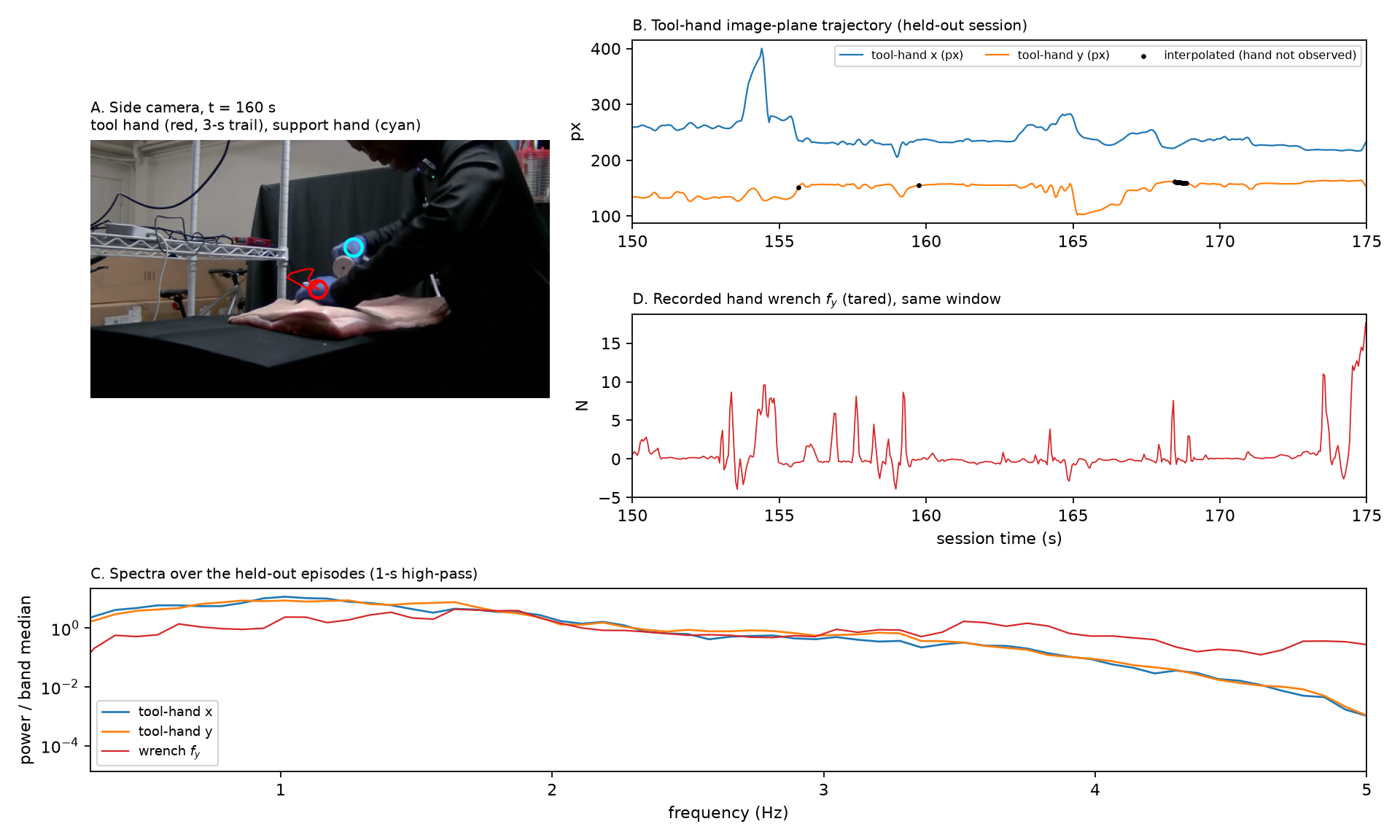}
\caption{The action channel. \textbf{A:} side-camera frame with the tracked tool hand (red)
and support hand (cyan). \textbf{B:} the tool-hand trajectory over 25\,s of the held-out
session; black dots mark interpolated samples. \textbf{C:} spectra of the hand track and of
the recorded force over the held-out episodes; the stroke rhythm is carried by the hand track.
\textbf{D:} the recorded force $f_y$ over the same window.}
\label{fig:hand}
\end{figure}

A demonstration dataset needs an \emph{action}: the quantity the demonstrator controls and a
robot would reproduce. In liposuction that is the cannula's motion, 1--1.7\,Hz aspiration
strokes of a few centimetres interleaved with repositioning. The tip is under the skin, but
the hand that drives it is in view of the side camera in every episode. Version~0 therefore
tracks that hand and records its image-plane trajectory as the action, with the next-step
position packaged in the chunked, state-relative convention that GR00T, ACT and $\pi_0$
expect (Fig.~\ref{fig:hand}).

The tracker resolves the \emph{tool} hand and the \emph{support} hand in every frame,
flags the frames in which the tool hand is hidden behind the sleeve or the instrument, and is
checked per session against overlay frames (Table~\ref{tab:track}). This channel is captured
by a model; the measured handle pose in Table~\ref{tab:sensing} upgrades it to metric units.
Method details are available to licensees.

\section{The dataset}
\label{sec:dataset}

\subsection{Sessions and scale}
\begin{figure}[t]
\centering
\includegraphics[width=0.85\linewidth]{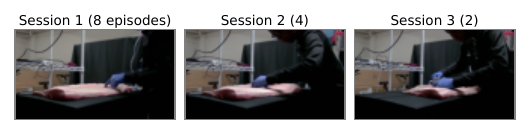}
\caption{The three capture sessions, side-camera view; episode counts in parentheses.}
\label{fig:sessions}
\end{figure}

Three sessions of 22, 10 and 6 minutes, segmented automatically into 14 episodes by the
suction-pressure signal (an episode is a stretch of active aspiration), for 35.6 minutes of
in-episode content at 20\,Hz (Fig.~\ref{fig:sessions}). Session~3 is reserved as the test
session for the proof of concept (Sec.~\ref{sec:poc}).

\subsection{What a record contains}
\begin{figure}[t]
\centering
\includegraphics[width=\linewidth]{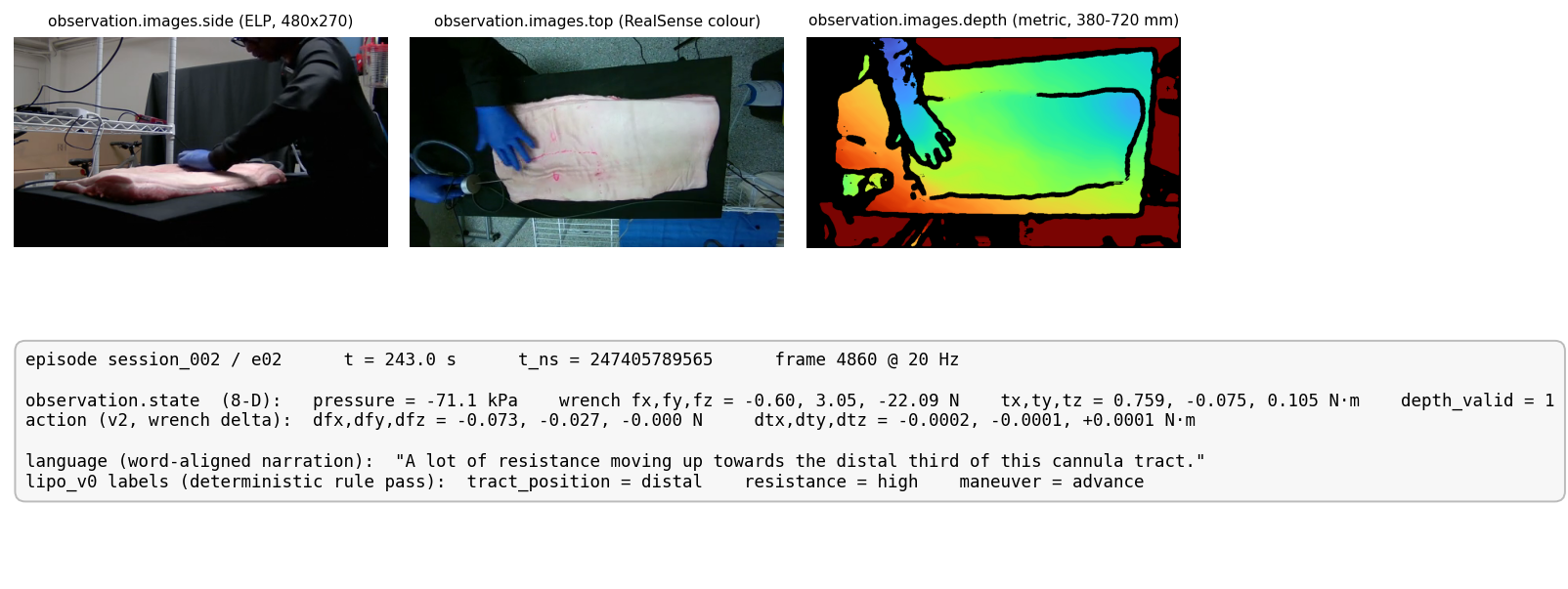}
\caption{One record with its images (session~2, $t=243.0$\,s): side RGB, top-down colour,
metric depth, with the frame's telemetry, narration and labels.}
\label{fig:record}
\end{figure}

Every frame is one row of a LeRobot~\cite{lerobot} dataset: three images, a 15-dimensional
state, an 18-dimensional action, and a language string, with per-frame labels alongside
(Table~\ref{tab:scale}). Table~\ref{tab:rows} shows six consecutive records exactly as a
model receives them, and Figure~\ref{fig:record} shows one with its images.

\begin{table}[h]
\centering\small
\caption{\dsname{} version~0 at a glance.}
\label{tab:scale}
\begin{tabular}{lR{13.2cm}}
\toprule
Field & Content\\
\midrule
Images & side RGB; top-down colour; metric depth (380--720\,mm window), all $480\times270$ at 20\,Hz\\
State (15-D) & suction pressure; 6-axis wrench; depth-valid flag; tool-hand $xy$ and velocity; support-hand $xy$; hand-valid flag\\
Action (18-D) & next-step tool-hand $xy$ (primary) and its delta; next-step support-hand $xy$; next-step wrench and its delta\\
Language & the surgeon's utterance at that frame (word-aligned), 356 utterances\\
Labels & 9 \texttt{lipo\_v0} fields per frame; 69.4\% of frames carry at least one\\
Raw streams & 467\,Hz wrench, 100\,Hz pressure, 48\,kHz audio, RealSense bags, ProRes side video, all with timing reports\\
Packaging & LeRobot v2 parquet + H.264 video + GR00T modality config; Open-H-compatible fields\\
\bottomrule
\end{tabular}
\end{table}

\input{tables_v4_rows.tex}

\subsection{The narration and its labels}
\label{sec:labels}
The surgeon narrates in a say-before-act cadence: announce the target (region, tract,
plane), act, then report what was felt and any assessment or safety reasoning. Verbatim,
from session~2:
\begin{quote}\small
``Inserting the cannula, tissue firm in the midsection of the hypogastric abdomen,
advancing, more friction\ldots'' \quad ``I'm in the deep fat.'' \quad
``Moving back to the proximal fat --- difficult here --- easy now.'' \quad
``A lot of resistance moving up towards the distal third of this cannula tract.'' \quad
``Don't want to make an aesthetic error by being too close to the skin.'' \quad
``The hypogastrium pinch test is symmetrical and equal.''
\end{quote}
Audio is transcribed with word times (WhisperX), aligned to the sensor timeline, and passed
through a deterministic rule set (\texttt{lipo\_v0}) into nine frame-level fields. Every
count in this document is auditable back to verbatim text. Table~\ref{tab:events} gives the
event families; at frame level 69.4\% of the 42{,}738 frames carry at least one label, with
clinically sensible joint structure (for example, \emph{superficial} $\times$
\emph{high-resistance} frames coincide with narrated caution about skin proximity). No
comparable frame-level record of surgical reasoning exists for any aesthetic procedure.

\begin{table}[h]
\centering\small
\caption{Narrated event tags, utterance level, all sessions (35.6 min): 356 utterances, 95\%
carrying at least one tag, about one tagged maneuver every 15 seconds.}
\label{tab:events}
\begin{tabular}{llc}
\toprule
Event family (count) & Top tags & Rate\\
\midrule
Maneuver (146) & reposition-lateral 42, move-back 22, stroke 20, advance 15, insert 14 & 4.1/min\\
Resistance (91) & high / low / increasing / decreasing & 2.6/min\\
Tract position (76) & proximal / middle / distal & 2.1/min\\
Difficulty (67) & hard / easy / moderate & 1.9/min\\
Tissue quality (60) & fibrous / firm / yielding / thin & 1.7/min\\
Plane (59) & deep / superficial / middle & 1.7/min\\
Safety (45) & error-risk 18, control 16, injury-risk 6, caution 5 & 1.3/min\\
Assessment (39) & end-point reached 23, aesthetic concern 7, pinch test 5 & 1.1/min\\
\bottomrule
\end{tabular}
\end{table}

\begin{figure}[t]
\centering
\includegraphics[width=0.9\linewidth]{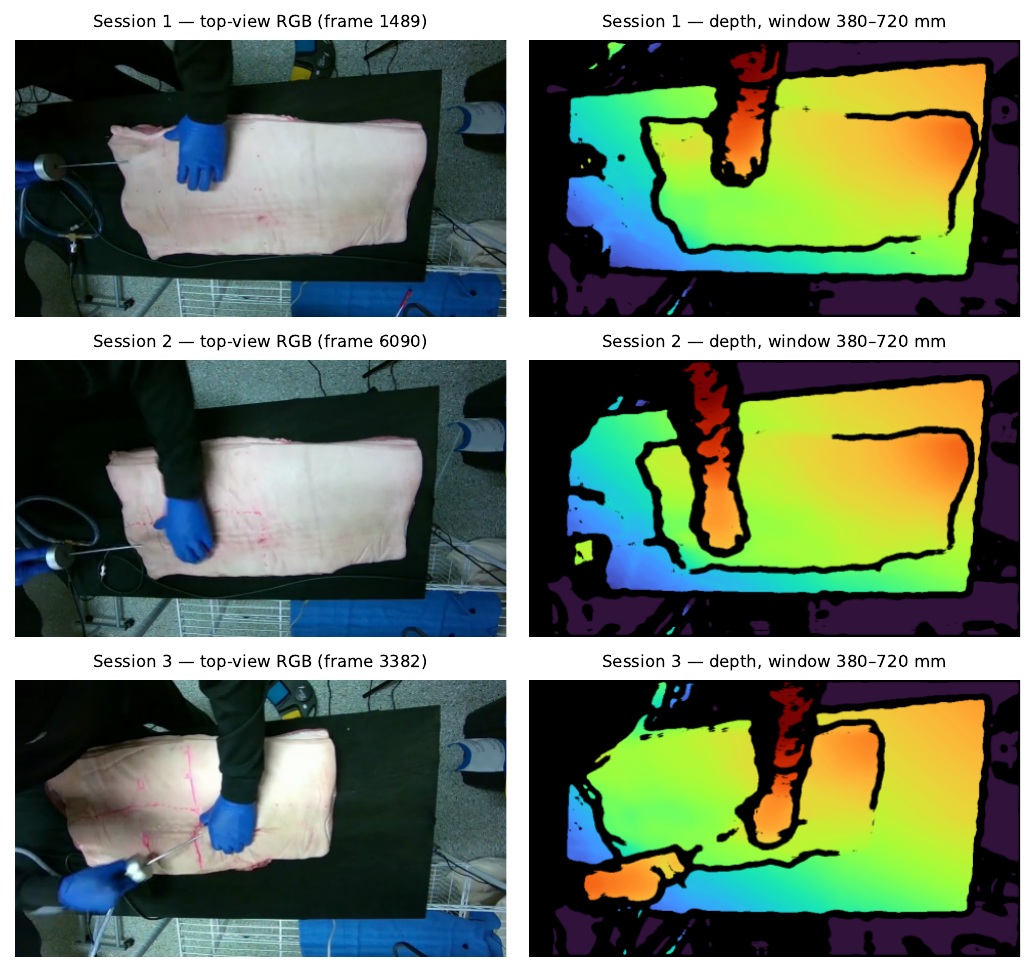}
\caption{The top-down channel per session: RealSense colour (left) and metric depth in the
fixed 380--720\,mm window (right).}
\label{fig:depth}
\end{figure}

\section{Data quality assurance}
\label{sec:quality}
A curated dataset is worth what its quality assurance is worth, so every channel in
\dsname{} ships with a per-session characterization and every derived channel carries a
per-frame validity flag.

\textbf{The hand tracker.} Table~\ref{tab:track} reports, per session, how often each hand
was directly observed and the dominant stroke frequency of the tool-hand track. Frames in
which the tool hand is not directly observed are interpolated and flagged in the data, so a
model can weight or skip them.

\input{tables_v3_track_public.tex}

\textbf{The force channel.} The six-axis force/torque stream is characterized against the
tool-hand track in every session and provided as state; the tool-hand trajectory is the
action target (Sec.~\ref{sec:action}). The next capture re-mounts and validates the sensor
before every session, which makes the force channel a second action target
(Sec.~\ref{sec:next}).

\textbf{Per-session reports.} Timing health, sensor drift, force--motion coherence and
depth validity are characterized for every session and delivered with the data; from the
next capture these checks run between episodes as gates (Sec.~\ref{sec:next}).

\section{Proof of concept: what a buyer can do with the data}
\label{sec:poc}

\subsection{Fine-tuning a foundation policy on the corpus}
The dataset loads into NVIDIA Isaac GR00T N1.7 (3B parameters) with no custom code. Each
fine-tuning run keeps the language and vision towers frozen and trains the projector and
action head for 10{,}000 steps (about one pass over the data), which takes
\vthreeKZeroWallMin{} minutes on one H100 GPU, under \$5 of compute. Every run is scored the
same way: on session~3, which is reserved as the test session and is never used
for training, the policy is shown the true observations at
every 16th frame and asked for the next 16 steps of tool-hand motion, and the error is
reported as mean absolute error in pixels of the side frame. Three seeds are reported as
mean $\pm$ sd, and the final checkpoint is the result. The full protocol, with reference
predictors scored on the same frames, is in Appendix~\ref{app:full}.

\subsection{What the runs show}
Table~\ref{tab:poc} lists the arms. Fine-tuning converges cleanly in every arm, the policy
learns the recorded sessions (in-sample error \vthreeKZeroInsampleMAE{}\,px), and the three
seeds agree to a tenth of a pixel, so the numbers are reproducible run to run. The arms
also say what the corpus responds to at this scale: adding the top-down colour and depth
views lowers the error on the unseen session by \vthreeKTwoGainPct{}, training twice as long
lowers it by \vthreeKZeroLGainPct{}, and the narration channel is neutral on this kinematic
target. The probes below ask where further gains on an unseen session come from.

\begin{table}[H]
\centering\small
\caption{Proof-of-concept arms: GR00T N1.7 fine-tuned on sessions~1+2 and scored on
session~3 (tool-hand MAE, px; lower is better).}
\label{tab:poc}
\begin{tabular}{llr}
\toprule
Arm & What it tests & MAE (px) \\
\midrule
Side RGB, 3 seeds & the base recipe & \vthreeKZeroSeedsMeanMAE{} $\pm$ \vthreeKZeroSeedsSdMAE{} \\
Side RGB, no narration & the language channel & \vthreeKZeronlHeldoutMAE{} \\
Side + top colour + depth & extra views & \vthreeKTwoHeldoutMAE{} \\
Side RGB, 20{,}000 steps & a longer schedule & \vthreeKZeroLHeldoutMAE{} \\
Both hands as target & a bimanual target (pooled) & \vthreeKZerobHeldoutMAE{} \\
\bottomrule
\end{tabular}
\end{table}

\subsection{Does more data help? Scaling and adaptation probes}
\label{sec:scaling}
\begin{figure}[t]
\centering
\includegraphics[width=\linewidth]{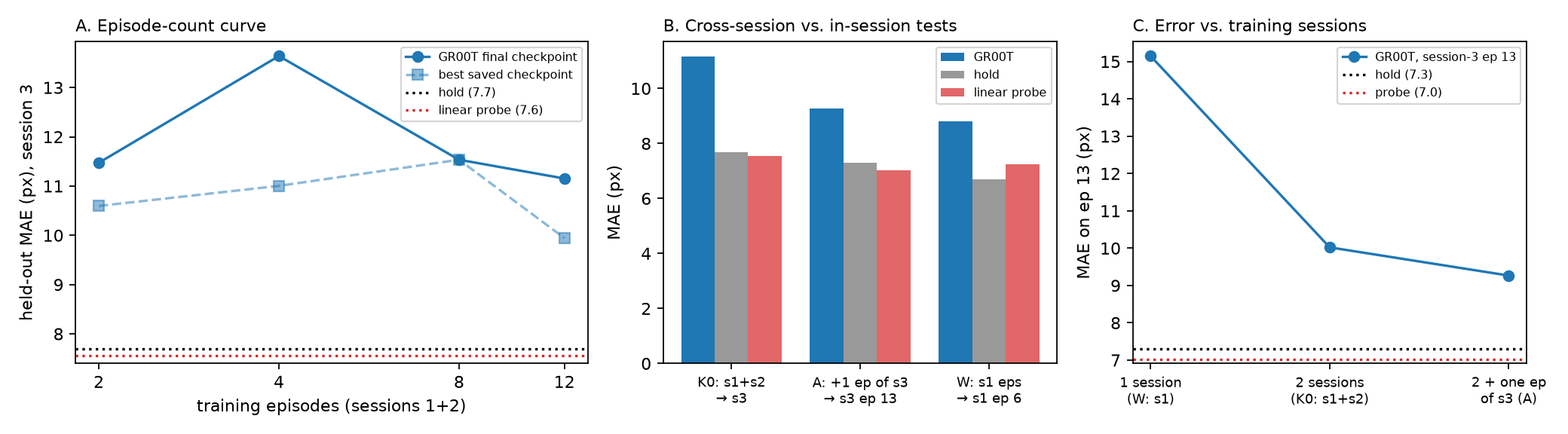}
\caption{Scaling and adaptation probes. \textbf{A:} held-out error on session~3 against the
number of training episodes drawn from sessions~1+2, with the baselines. \textbf{B:} the
cross-session test next to two in-session tests: one episode of the target session added to
training (A), and an unseen episode of a trained session (W). \textbf{C:} error on one
held-out episode against the number of training sessions.}
\label{fig:scaling}
\end{figure}

The question that matters for the next capture is where further gains come from: more
episodes, or more sessions? Three probes on the existing episodes answer it
(Fig.~\ref{fig:scaling}, Table~\ref{tab:v4scaling}); each uses its own training and test
episodes, and all were specified before they were run.

\input{tables_v4_scaling.tex}

\begin{itemize}\setlength\itemsep{0.1em}
  \item \textbf{Every new session helps.} Trained on session~1 alone the policy scores
  \vthreeScalOneSessionSThree{}\,px on session~3; on sessions~1+2, \vthreeScalTwoSessionsSThree{}\,px,
  \vthreeSessionGainPct{} lower. Adding a single 2.6-minute episode of session~3 to training
  lowers the error on that session's other episode from \vthreeScalEpThirteenTwoSessions{} to
  \vthreeScalEpThirteenTwoPlusPartial{}\,px (Fig.~\ref{fig:scaling}C).
  \item \textbf{The gain is specific to new sessions.} With the sessions held fixed,
  varying the number of training episodes from two to twelve leaves the error on session~3
  within a narrow band (Table~\ref{tab:v4scaling}), so the return on the next capture comes
  from session diversity.
\end{itemize}
The reading for the next capture is direct. The lever is session diversity, which in this
specimen means fat region and composition (Table~\ref{tab:tissue}), and the corpus is the
instrument that measures the return on each added session, using the same protocol.

\section{What the next capture changes}
\label{sec:next}
\begin{itemize}\setlength\itemsep{0.1em}
  \item \textbf{Many short sessions across fat regions.} One session per region and
  composition (abdomen, flank, thigh; fibrous, firm, yielding; superficial and deep planes),
  targeting at least ten sessions and about 100 episodes, the scale at which GR00T-class
  adaptation is expected to show clear signal. The probes of Section~\ref{sec:scaling} put
  the value in new sessions.
  \item \textbf{Sensors in place of models.} Four upgrades change what the data contains
  (Table~\ref{tab:sensing}). A 6-DoF pose sensor on the handle turns the 2-D tracked hand into
  a measured action in metric units for every frame (tier A). The force/torque sensor is
  re-mounted and validated before every session, which makes the force channel a second
  action target (tier A). Ultrasound imaging of the fat layer per region
  adds the state of the tissue the tip is working in (tier B). Tactile
  sensing on the other hand records what the surgeon feels when palpating (tier B). Contact and
  suction-line sensing, an egocentric view and ground-truth tracking complete the rig (tiers
  B--C). Tier A is a \$2--4k increment on the rig we own; tier B adds \$8--15k; tier C adds
  \$15--40k.
  \item \textbf{Quality gates at the table.} Timing, sensor drift, force--motion coherence
  and depth validity are checked between episodes, in seconds, so every episode that enters
  the corpus has passed them.
  \item \textbf{Then cadaveric tissue and a second expert}, to ground anatomical fidelity
  and to measure inter-surgeon transfer.
  \item \textbf{Throughout}, Open-H/LeRobot-compatible packaging with per-sensor provenance
  and QA reports, which is what keeps the corpus licensable and regulator-legible as it
  grows.
\end{itemize}

\section{Conclusion}
\dsname{} version~0 is a working expert dataset. A master liposuction surgeon's hands and
reasoning are captured together: 14 narrated episodes with synchronized suction pressure,
hand force/torque, top-down colour and depth video, side video and word-aligned narration
compiled into labels, packaged so that a state-of-the-art robot foundation model fine-tunes
on it in \vthreeKZeroWallMin{} minutes with no custom code.

The same experiments show where the value grows. Adding one session to training lowered the
error on an unseen session by \vthreeSessionGainPct{}, and the sessions differed in fat
region and composition, so the next capture is many short sessions across regions and tissue
types. The sensing plan then replaces the channels a model estimates today with measured
ones: a 6-DoF handle pose, a validated force channel, ultrasound imaging of the fat layer and
palpation sensing, all without changing the data format. More varied sessions and better
sensors are the two investments that turn this proof of concept into the expert liposuction
corpus that surgical robot models currently lack.

\clearpage
\section*{Reproducibility and provenance}
\begin{itemize}\setlength\itemsep{0.15em}
  \item \textbf{Numbers.} Every number in this document is generated from the recorded
  artifacts and mapped claim by claim in an internal provenance record.
  \item \textbf{Protocol.} The evaluation protocol is specified in Section~\ref{sec:poc} and
  Appendix~\ref{app:full}, and every future release is scored by it.
  \item \textbf{Access.} The dataset, its per-session quality reports and the tooling are
  available to licensees under agreement.
\end{itemize}

% appendix tables are placed where they are written, in order
\makeatletter
\renewenvironment{table}[1][t]{\@float{table}[H]}{\end@float}
\renewenvironment{table*}[1][t]{\begin{table}}{\end{table}}
\makeatother

\appendix
\section{Full evaluation tables}
\label{app:full}
The tables below give every arm with the reference predictors scored on the same frames
(hold: predict no motion; constant velocity; training mean; and a linear probe fit on the
training episodes), the observed-frames-only and per-axis breakdown, the error at every
saved checkpoint, and the scaling probes with their reference predictors.
\input{tables_v3_experiments.tex}
\input{tables_v3_masked.tex}
\input{tables_v3_curve.tex}
\input{tables_v4_scaling_full.tex}

\end{document}

%% file: v3_numbers.tex
\newcommand{\vthreeKZeroInsampleMAE}{8.13}

\newcommand{\vthreeKZeroWallMin}{36}

\newcommand{\vthreeKZeronlHeldoutMAE}{11.18}

\newcommand{\vthreeKTwoHeldoutMAE}{10.67}

\newcommand{\vthreeKZerobHeldoutMAE}{11.40}

\newcommand{\vthreeKZeroLHeldoutMAE}{10.73}

\newcommand{\vthreeKZeroSeedsMeanMAE}{11.20}
\newcommand{\vthreeKZeroSeedsSdMAE}{0.07}

\newcommand{\vthreeKTwoGainPct}{5\%}
\newcommand{\vthreeKZeroLGainPct}{4\%}

\newcommand{\vthreeTrackValidSOne}{90\%}

\newcommand{\vthreeTrackValidSThree}{82\%}

%% file: v3_scaling_numbers.tex
\newcommand{\vthreeSessionGainPct}{18\%}
\newcommand{\vthreeScalOneSessionSThree}{13.56}
\newcommand{\vthreeScalTwoSessionsSThree}{11.15}

\newcommand{\vthreeScalEpThirteenTwoSessions}{10.02}
\newcommand{\vthreeScalEpThirteenTwoPlusPartial}{9.27}

%% file: tables_v4_tissue.tex
\begin{table}[H]
\centering
\caption{What the surgeon said about the tissue in each session, as the share of that
session's frames carrying each narrated tag (\texttt{lipo\_v0}; top three values per field).
The sessions worked different regions of the specimen and read as different material.}
\label{tab:tissue}
\footnotesize
\begin{tabular}{lp{3.2cm}p{3.0cm}p{3.0cm}p{3.0cm}}
\toprule
Session & Tissue quality & Plane & Resistance & Difficulty \\
\midrule
Session 1 & fibrous 7\%, yielding 4\%, firm 3\% & deep 13\%, superficial 4\%, middle 2\% & low 6\%, high 5\%, increasing 4\% & hard 8\%, easy 6\%, moderate 3\% \\
Session 2 & firm 9\%, yielding 1\%, thin 1\% & superficial 17\%, deep 7\%, middle 3\% & high 25\%, low 9\%, decreasing 6\% & hard 27\%, moderate 5\%, easy 2\% \\
Session 3 & yielding 13\%, fibrous 6\%, firm 2\% & superficial 1\% & low 6\%, decreasing 5\%, increasing 2\% & easy 6\%, hard 5\% \\
\bottomrule
\end{tabular}
\end{table}

%% file: tables_v4_rows.tex
\begin{table}[H]
\centering
\caption{Six consecutive records, one second apart, exactly as a model receives them
(session~2, episode~9, from $t = 243$\,s). Every row carries the three synchronized
images (side RGB, top-down colour, depth; not shown), the suction pressure, the recorded
wrench (three force axes shown), the tool-hand and support-hand positions in the side
frame, whether the tool hand was directly observed (\good) or interpolated (\bad), the
next-step tool-hand target, the surgeon's words at that instant, and the tags compiled
from them.}
\label{tab:rows}
\scriptsize
\setlength{\tabcolsep}{1.5pt}
\begin{tabular}{rrlllcl>{\raggedright\arraybackslash}p{3.0cm}>{\raggedright\arraybackslash}p{2.9cm}}
\toprule
$t$ (s) & $P$ (kPa) & $f_x, f_y, f_z$ (N) & tool hand (px) & support (px) & obs. & next target & narration & \texttt{lipo\_v0} tags \\
\midrule
243.0 & -71.2 & -0.5, +2.7, -22.1 & (263, 145) & (316, 132) & \bad & (278, 142) & \emph{A lot of resistance moving up towards the distal thi\ldots} & tract\_position=distal, resistance=high, maneuver=advance \\
244.0 & -70.8 & -0.1, +2.6, -22.1 & (297, 139) & (338, 139) & \good & (297, 139) & \emph{A lot of resistance moving up towards the distal thi\ldots} & tract\_position=distal, resistance=high, maneuver=advance \\
245.0 & -71.1 & +0.1, +2.4, -22.0 & (229, 147) & (260, 142) & \good & (231, 146) & \emph{A lot of resistance moving up towards the distal thi\ldots} & tract\_position=distal, resistance=high, maneuver=advance \\
246.0 & -71.2 & -0.1, +2.4, -22.3 & (234, 150) & (266, 147) & \good & (235, 149) & \emph{A lot of resistance moving up towards the distal thi\ldots} & tract\_position=distal, resistance=high, maneuver=advance \\
247.0 & -71.2 & -0.5, +3.0, -22.1 & (232, 150) & (260, 146) & \good & (228, 151) & \emph{A lot of resistance moving up towards the distal thi\ldots} & tract\_position=distal, resistance=high, maneuver=advance \\
248.0 & -71.8 & -0.7, +3.0, -22.1 & (228, 155) & (231, 155) & \bad & (228, 155) & \emph{A lot of resistance moving up towards the distal thi\ldots} & tract\_position=distal, resistance=high, maneuver=advance \\
\bottomrule
\end{tabular}
\end{table}

%% file: tables_v3_track_public.tex
\begin{table}[H]
\centering
\caption{Side-camera hand tracking per session: frames, share of frames in which the tool
hand and the support hand were directly observed (the rest are interpolated and flagged), and
the dominant stroke frequency of the tool-hand track ($x$ / $y$, 0.6--4\,Hz band).}
\label{tab:track}
\footnotesize
\begin{tabular}{lrrrr}
\toprule
Session & Frames & Tool hand observed & Support hand observed & Stroke peak (Hz) \\
\midrule
Session 1 & 25,649 & 90\% & 86\% & 1.00 / 1.00 \\
Session 2 & 12,304 & 78\% & 79\% & 0.88 / 1.12 \\
Session 3 & 6,686 & 82\% & 84\% & 1.00 / 0.88 \\
\bottomrule
\end{tabular}
\end{table}

%% file: tables_v4_scaling.tex
\begin{table}[H]
\centering
\caption{Scaling and adaptation probes on the hand target (MAE, px; same protocol as the
proof-of-concept arms). Each probe has its own training and test episodes; the full
comparison with reference predictors is in Appendix~\ref{app:full}.}
\label{tab:v4scaling}
\footnotesize
\begin{tabular}{llrr}
\toprule
Arm & Split & Train eps & MAE (px) \\
\midrule
W & within-session: s1 eps 0--5,7 $\rightarrow$ ep 6 & 7 & 8.82 \\
A & adaptation: all but ep 13 $\rightarrow$ ep 13 & 13 & 9.27 \\
N2 & 2 episodes (0, 8) $\rightarrow$ session 3 & 2 & 11.47 \\
N4 & 4 episodes $\rightarrow$ session 3 & 4 & 13.64 \\
N8 & 8 episodes $\rightarrow$ session 3 & 8 & 11.53 \\
K0 & 12 episodes (K0) $\rightarrow$ session 3 & 12 & 11.15 \\
\bottomrule
\end{tabular}
\end{table}

%% file: tables_v3_experiments.tex
\begin{table*}[t]
\centering
\caption{Open-loop evaluation with a \emph{matched} protocol. At every 16th frame of a
trajectory the policy (or baseline) emits a 16-step chunk from ground-truth observations;
the chunk is scored against the recorded future (MAE/MSE over all channels and steps, raw
units). Whole episodes are scored. Held-out = Session~3 (episodes 12--13), never seen in
training; in-sample = episodes 0--2. The linear AR probe is a ridge regression fit on the
training episodes only.}
\label{tab:v3exp}
\footnotesize
\begin{tabular}{lrrrrl}
\toprule
 & \multicolumn{2}{c}{Held-out} & \multicolumn{2}{c}{In-sample} & \\
Policy / baseline & MAE & MSE & MAE & MSE & Model \\
\midrule
\multicolumn{6}{l}{\emph{Target: tool-hand trajectory $h_{t+1:t+16}$ (px, side camera), predicted relative to $h_t$}} \\
Hold $h_t$ (zero motion) & 7.70 & 297 & 7.50 & 202 & baseline \\
Constant velocity & 13.97 & 1233 & 13.70 & 748 & baseline \\
Training mean & 33.19 & 1888 & 24.93 & 1005 & baseline \\
Linear AR probe (ridge, $K$=8) & 7.55 & 257 & 7.02 & 175 & baseline \\
\textbf{K0: side RGB} (3 seeds, mean$\pm$sd) & \textbf{11.20$\pm$0.07} & 399 & 8.14$\pm$0.01 & 190 & GR00T N1.7 \\
\quad seed 0 & 11.15 & 409 & 8.13 & 196 & \\
\quad seed 1 & 11.17 & 408 & 8.14 & 186 & \\
\quad seed 2 & 11.29 & 381 & 8.15 & 188 & \\
K0nl: side RGB, no narration & 11.18 & 419 & 8.87 & 213 & GR00T N1.7 \\
K2: side RGB + top colour + depth & 10.67 & 362 & 7.80 & 177 & GR00T N1.7 \\
\midrule
\multicolumn{6}{l}{\emph{Target: both hands $[h, s]_{t+1:t+16}$ (px), pooled}} \\
Hold (zero motion) & 7.68 & 316 & 6.97 & 207 & baseline \\
Linear AR probe & 7.58 & 255 & 6.75 & 172 & baseline \\
K0b: side RGB, bimanual & 11.40 & 405 & 7.96 & 185 & GR00T N1.7 \\
\midrule
\multicolumn{6}{l}{\emph{Control target: recorded hand wrench $w_{t+1:t+16}$ (N / N\,m), predicted relative to $w_t$}} \\
Hold $w_t$ & 0.285 & 0.99 & 0.186 & 0.34 & baseline \\
Constant velocity & 1.111 & 23.64 & 0.702 & 7.34 & baseline \\
Training mean & 0.344 & 0.78 & 0.355 & 0.51 & baseline \\
K0w: side RGB, wrench target & 0.327 & 0.92 & 0.206 & 0.29 & GR00T N1.7 \\
\bottomrule
\end{tabular}
\end{table*}

%% file: tables_v3_masked.tex
\begin{table}[t]
\centering
\caption{Held-out hand-target MAE (px) on all frames; on frames whose target position was
observed by the tracker (\emph{observed}; 2590 of 3171 frames); per image axis; and split by
whether the hand actually travels more than 5\,px within the 16-step chunk (\emph{moving},
74\% of frames) or not (\emph{still}). Same frames and protocol for every row.}
\label{tab:v3masked}
\footnotesize
\begin{tabular}{lrrrrrrl}
\toprule
Policy / baseline & All & Observed & $x$ & $y$ & Moving & Still & Model \\
\midrule
Hold $h_t$ & 7.70 & 7.97 & 10.41 & 4.98 & 10.10 & 0.96 & baseline \\
Constant velocity & 13.97 & 15.47 & 17.56 & 10.37 & 18.07 & 2.38 & baseline \\
Training mean & 33.19 & 32.63 & 50.82 & 15.55 & 28.59 & 46.07 & baseline \\
Linear AR probe & 7.55 & 7.77 & 10.32 & 4.78 & 9.35 & 2.50 & baseline \\
\textbf{K0: side RGB} (3 seeds, mean) & 11.20 & 10.95 & 16.08 & 6.33 & 11.97 & 8.97 & GR00T N1.7 \\
K0nl (no narration) & 11.18 & 11.18 & 15.74 & 6.63 & 12.61 & 7.10 & GR00T N1.7 \\
K2 (3 views) & 10.67 & 10.75 & 14.91 & 6.43 & 11.58 & 8.12 & GR00T N1.7 \\
K0b (bimanual; hand part) & 11.35 & 11.15 & 16.11 & 6.59 & 11.80 & 10.17 & GR00T N1.7 \\
\bottomrule
\end{tabular}
\end{table}

%% file: tables_v3_curve.tex
\begin{table}[t]
\centering
\caption{Held-out MAE of every arm at each saved checkpoint (diagnostic; the final
checkpoint is the pre-registered result). Units: px for hand targets, N / N\,m for K0w. K0L
trains for 20k steps, so its final column is step 20k; ``--'' = no checkpoint at that step.}
\label{tab:v3curve}
\footnotesize
\begin{tabular}{lrrrrrrrr}
\toprule
Arm & step 2500 & step 5000 & step 7500 & step 10000 & step 12500 & step 15000 & step 17500 & final \\
\midrule
K0 & 14.08 & 13.63 & 9.94 & -- & -- & -- & -- & 11.15 \\
K0s1 & 14.01 & 13.84 & 10.59 & -- & -- & -- & -- & 11.17 \\
K0s2 & 16.77 & 12.90 & 10.16 & -- & -- & -- & -- & 11.29 \\
K0nl & 13.01 & 12.36 & 10.24 & -- & -- & -- & -- & 11.18 \\
K2 & 12.57 & 13.53 & 9.32 & -- & -- & -- & -- & 10.67 \\
K0b & 14.43 & 13.67 & 10.23 & -- & -- & -- & -- & 11.40 \\
K0w & 0.834 & 0.516 & 0.456 & -- & -- & -- & -- & 0.327 \\
K0L & 16.19 & 14.32 & 12.24 & 13.61 & 11.33 & 9.95 & 11.04 & 10.73 \\
\bottomrule
\end{tabular}
\end{table}

%% file: tables_v4_scaling_full.tex
\begin{table}[t]
\centering
\caption{Scaling and adaptation probes on the hand target (held-out MAE, px; same protocol as
Table~\ref{tab:v3exp}), with the reference predictors fit on the same training episodes and
scored on the same test episodes (hold and a linear probe). ``Best ckpt'' is the lowest of the
saved checkpoints (diagnostic).}
\label{tab:v4scalingfull}
\footnotesize
\begin{tabular}{llrrrrr}
\toprule
Arm & Split & Train eps & MAE & Hold & Probe & Best ckpt \\
\midrule
W & within-session: s1 eps 0--5,7 $\rightarrow$ ep 6 & 7 & 8.82 & 6.71 & 7.25 & 8.53 \\
A & adaptation: all but ep 13 $\rightarrow$ ep 13 & 13 & 9.27 & 7.30 & 7.02 & 9.27 \\
N2 & 2 episodes (0, 8) $\rightarrow$ session 3 & 2 & 11.47 & 7.70 & 7.49 & 10.60 \\
N4 & 4 episodes $\rightarrow$ session 3 & 4 & 13.64 & 7.70 & 7.45 & 11.00 \\
N8 & 8 episodes $\rightarrow$ session 3 & 8 & 11.53 & 7.70 & 7.67 & 11.53 \\
K0 & 12 episodes (K0) $\rightarrow$ session 3 & 12 & 11.15 & 7.70 & 7.55 & 9.94 \\
\bottomrule
\end{tabular}
\end{table}